%% file: template.tex
\documentclass{article}

\usepackage{arxiv}
\usepackage{amsmath}
\usepackage{mathtools}
\usepackage[utf8]{inputenc} % allow utf-8 input
\usepackage[T1]{fontenc}    % use 8-bit T1 fonts
\usepackage{hyperref}       % hyperlinks
\usepackage{url}            % simple URL typesetting
\usepackage{booktabs}       % professional-quality tables
\usepackage{amsfonts}       % blackboard math symbols
\usepackage{nicefrac}       % compact symbols for 1/2, etc.
\usepackage{microtype}      % microtypography
\usepackage{amsmath}
\usepackage{cleveref}       % smart cross-referencing
\usepackage{lipsum}         % Can be removed after putting your text content
\usepackage{graphicx}
\usepackage{natbib}
\usepackage{doi}
\usepackage{wrapfig}

\newcommand{\etal}{\textit{et al}. }

\title{Two-Stream UNET Networks for Semantic Segmentation in Medical Images}

\author{ Xin Chen, Ke Ding \\
	Intel Corp.\\
	2200 Mission College Blvd\\
	Santa Clara, CA 95054 \\
	\texttt{(xin.chen, ke.ding)@intel.com} \\
}

\renewcommand{\shorttitle}{\textit{arXiv} Template}

\hypersetup{
pdftitle={A template for the arxiv style},
pdfsubject={q-bio.NC, q-bio.QM},
pdfauthor={David S.~Hippocampus, Elias D.~Striatum},
pdfkeywords={First keyword, Second keyword, More},
}

\begin{document}
\maketitle

\begin{abstract}
	%   We investigate architectures of discriminatively trained deep convolutional neural Networks (CNNs) for semantic segmentation in medical images. The challenge is to capture the complementary information on appearance from still frames and motion be-tween frames. We also aim to generalise the best performing hand-crafted featureswithin a data-driven learning framework.Our contribution is three-fold. First, we propose a two-stream ConvNet architec-ture which incorporates spatial and temporal networks. Second, we demonstrate that a ConvNet trained on multi-frame dense optical flow is able to achieve very good performance in spite of limited training data. Finally, we show that multi-task learning, applied to two different action classification datasets, can be used toincrease the amount of training data and improve the performance on both. Ourarchitecture is trained and evaluated on the standard video actions benchmarks ofUCF-101 and HMDB-51, where it is competitive with the state of the art. It alsoexceeds by a 
	
	% We present two-stream U-net architecture for semantic segmentation in medical images. 
	Recent advances of semantic image segmentation greatly benefit from deeper and larger Convolutional Neural Network~(CNN) models. Compared to image segmentation in the wild, properties of both medical images themselves and of existing medical datasets hinder training deeper and larger models because of overfitting. To this end, we propose a novel two-stream UNET architecture for automatic end-to-end medical image segmentation, in which intensity value and gradient vector flow~(GVF) are two inputs for each stream, respectively. We demonstrate that two-stream CNNs with more low-level features greatly benefit semantic segmentation for imperfect medical image  datasets. Our proposed two-stream networks are trained and evaluated on the popular medical image segmentation benchmarks, and the results are competitive with the state of the art. The code will be released soon.
	% It alsoexceeds by a large margin previous attempts to use deep ne for video classifica-tion

	% proprieties of medical images and status of datasets, we  of Our contribution is three-fold. First, we propose a two-stream CNNs a It consists of an LSTR encoderthat dynamically leverages coarse-scale historical information from an extendedtemporal window (e.g., 2048 frames spanning of up to 8 minutes), together withan LSTR decoder that focuses on a short time window (e.g., 32 frames spanning8 seconds) to model the fine-scale characteristics of the data. Compared to priorwork, LSTR provides an effective and efficient method to model long videos withfewer heuristics, which is validated by extensive empirical analysis. LSTR achievesstate-of-the-art performance on three standard online action detection benchmarks,THUMOS’14, TVSeries, and HACS Segment. Code has been made available at:https://xumingze0308.github.io/projects/lstr.
\end{abstract}

\input{sections/introduction}
\input{sections/relatedwork}
\input{sections/twostream}
\input{sections/tech_approach}
\input{sections/experiment}
\input{sections/evaluation}

\input{sections/conclusion}

\bibliographystyle{unsrt}
\bibliography{references}  %%% Uncomment this line and comment out the ``thebibliography'' section below to use the external .bib file (using bibtex) .





% \end{thebibliography}

\end{document}

%% file: sections/introduction.tex
\section{Introduction}
\label{sec_intr}

As an important subfield of image segmentation, medical image segmentation is a challenging task which has received a significant amount of attention in both academia and industry~\cite{pham2000current,Lei2022MedicalIS,duncan2000medical,sharma2010automated,ramesh2021review,tajbakhsh2016convolutional,smistad2015medical,garcia2018survey,yu2021convolutional}. Since emerging deep CNNs~\cite{yan2015deep} and their application to image segmentation~\cite{long2015fully}, the segmentation task has made a remarkable progress in recent years, which benefits from improving visual representations with deeper and larger of CNNs models~\cite{garcia2018survey,wang2018understanding,hao2020brief}. However, compared to images in the wild, the deeper and larger models are limited to medical segmentation because of the following challenges~\cite{hesamian2019deep,Malhotra2022DeepNN,liu2021review,tajbakhsh2020embracing}:
\begin{enumerate}
   \item \textbf{Properties of medical datasets.}~Size of medical image datasets are tiny. Privacy of patient information and labelling cost restrict to build large scale image datasets. Also, due to clinical applications, the number of categories is small, in general less than five and even only one. As a result, on average, size of the medical datasets is only one tenth or less than that of natural images.
    \item \textbf{Properties of medical images themselves.}~There are two challenges to train deeper models on medical images. First, medical images have similar intensities of pixels. Second, some factors of medical acquisition such as sampling artifacts, spatial aliasing, and some of the dedicated noise of modalities cause the indistinct and disconnected boundary's structures. %and is mostly occupied by either back or white not much texture information. Also the number of categories is small, in general less than five and even only one. ;
    % \item The number of categories is small, in general less than five and even only one. Compared to the image segmentation in the wild, the number of categories is large. Take the COCO dataset as an example~\cite{lin2014microsoft}, there are 80 object categories; from the perspective of
    
\end{enumerate}
% Semantic segmentation with deep learning techniques in images is a challenging task which has received a significant amount of attention in the research community~\cite{GARCIAGARCIA201841,lateef2019survey,hao2020brief,long2015fully}. Compared to image segmentation in the wild, medical image segmentation faces much challenges~\cite{Lei2022MedicalIS,duncan2000medical,liu2021review,Malhotra2022DeepNN}, for example 
Ronneberger~\etal~\cite{ronneberger2015u} develop a novel U-shape model of the encoder-decoder network architecture, named UNET, and it outperforms marginally on three small cell datasets. In the encoder-decoder networks, skip connections have proved that they play a key role to recover fine-grained details of the target images. Therefore, in order to aggregate fine-grained information efficiently and effectively from different layers, Zhou~\etal UNET++~\cite{zhou2019unetplusplus}, in which the topology of skip connections is redesigned. The advantage of the UNET++ architecture is that the models gradually aggregate features across the networks both horizontally and vertically. 

Given properties of the medical images, a well-know assumption is that much information that are fused appropriately is an effective way to improve segmentation results rather than heavily depending on recovering from one feature map~\cite{isensee2021nnu}. Take the EM segmentation task~\cite{ewald2012mammary} for example, in ~\cite{ronneberger2015u} and ~\cite{zhou2021towards}, besides accuracy improvement from better CNNs models side, the post-processing step, combining sliding window and voting rule on overlay areas, also contributes greatly to increase the accuracy. As a result, \emph{\textbf{a question arises: can we design a new architecture, in which multiple low-level features are fed into CNNs models and they can work on these multiple features?}}  

Inspired by two-stream hypothesis of human vision~\cite{simonyan2014two,ungerleider1994and} and state-of-the-art~(SOTA) of action recognition in videos using two-stream networks~\cite{simonyan2014two}, we propose a novel two-stream CNNs architecture for semantic segmentation in medical images, which incorporates spatial and vector field networks by introducing gradient vector flow~(GVF)~\cite{xu1997gradient} as the input of the vector field network, for semantic segmentation in medical images. The backbone of each stream network of our two-steam networks is basic UNET~\cite{ronneberger2015u} that is one of most popular CNNs architectures and is hypothesized that it is hard to beat if the corresponding pipeline is designed adequately~\cite{isensee2021nnu}. The final fusion layer takes convolution fusion with $1\times 1$ convolution kernel. 

\begin{figure}[!t]
    \centering
    \vspace{-1cm}
    \includegraphics[trim=0cm 3cm 0cm 3cm, clip, width=\columnwidth]{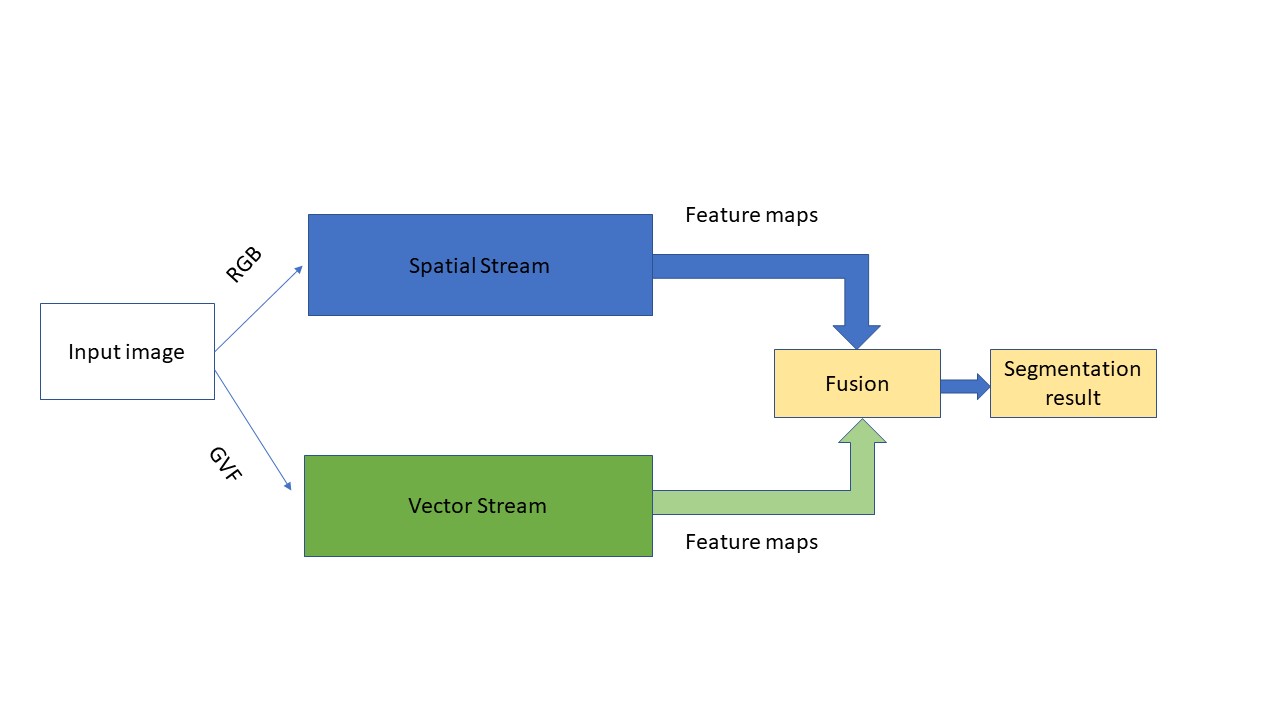}
   \vspace{-0.5cm}
    \caption{Visualization of our proposed two-stream UNET networks. We feed RGB~(3-channel) or gray-level~(1-channel) values into the spatial stream~(blue) and corresponding stacked GVF fields to the vector stream~(green). The backbone we take basic UNET model~\cite{ronneberger2015u}. The outputs of each stream is $64\times height\times width$. After the fusion layer, the output is $N\times height \times width$. N is number of classes. In the fusion layer, we took $1 \times 1$ convolutional kernel.}
    \label{fig:twostream}
\end{figure}

Our main contribution is to propose a novel two-stream UNET architecture to end-to-end learn for automatic segmentation in medical images. The two-stream networks consist of spatial and vector field networks, one is referred to as \textbf{\textit{spatial stream~(SS)}} and the other is referred to as \textbf{\textit{vector stream~(VS)}}. Our proposed model architecture is visualized in  Figure.~\ref{fig:twostream}. The input of the spatial stream is intensity value of the pixels, either RGB or grayscale-level, and the input of the vector field one is gradient vector flow~(GVF), 2-channel, as the input of the vector field network. GVF is considered as one of best low-level pixel-wise features and greatly improves original active contour model~(Snake)~\cite{kass1988snakes} acting as a new external force~\cite{xu1998snakes}. 

To the best knowledge, our proposed tow-stream UNET is first two-stream networks for end-to-end learn to segment in all kinds of medical images. Through introducing GVF to VS, the models to learn an object~(an pixel) how to moving to contours. Then our models reason positions of the contours. In turns, the two-stream networks successfully handle the segmentation task. Obviously, our work can be easily extended to other computer vision tasks, for example object detection, instance segmentation and so on. 

Our experimental results demonstrate that: (1) Our proposed two-stream UNET model marginally improves UNET and substantially outperforms other SOTA UNET variants. (2): The two-stream model attains competitive results compared to well-known networks such as vision transformer~(ViT) while requiring substantially fewer computational resources to train and inference. 

%% file: sections/relatedwork.tex
\section{Related work}
\label{sec_relatedwork}
There is an extensive literature on UNET and its variants, but here we mention just a few relevant papers on UNET architecture and and two-stream networks relative to image segmentation. Some surveys and relative works are found in~\cite{liu2020survey, siddique2021u,oktay2018attention,cciccek20163d}, and recently developed approaches are presented in the various challenges hold at annual MICCAI meetings~\footnote{\url{http://www.miccai.org/}}. 

In image segmentation task, most of existing two-stream networks aim to solve not enough labelling problems, for example different domain adaption~\cite{bermudez2018domain}, pseudo labeling with mutual attention network~\cite{min2019two}. Hu~\etal address a fully convolution two-stream network for interactive segmentation, which has a two-stream late fusion network~(TSLFN) and a multi-scale refining network~(MSRN). TSLFN predicts the foreground with a reduced resolution, and MSRN predicts the foreground at fill resolution. Then both are fused for final result. One input is RGB data and the other is the minimum Euclidean distance between pixels and the user click. The inherited problem of interactive approaches is that the predictions heavily reply on the quality of user clicks. Moreover, the whole process is involved by users. 

In UNET++~\cite{zhou2018unetplusplus,zhou2021towards}, besides redesigning skip connections which flexibly fuse features in decoders, deep supervision is utilized for improving model training. UNET++ models fuse image features across the network instead of the same-scale features in U-net. Combining with deep supervision,  NNET++ models outperform basic UNET averagely. However, considered low content of medical images, CNNs model represent less visual features of medical images than natural images.

We notice that in the past two year nnUNET~(no-new-UNet)~\cite{isensee2021nnu} has outperformed state-of-the-art~(SOTA) architectures in the Medical Decathlon Challenge~\cite{medicaldecathlon}, then has set new benchmarks in more datasets. First, nnUNET takes a well-known hypothesis that basic UNET is hard to beat. This hypothesis theoretically support why the basic UNET is well-suited for our two-stream networks for image segmentation. Second, nnUNET comprises of ten different datasets using an ensemble of the same U-Net architecture with an automated pipeline comprising of pre-processing, data augmentation and post-processing. nnUNET is time cost is experience in both training and inference because nnUNET requires more than one models in inference. Also, the nnUNET model is not good at small objects and noise edges because there are resize operations with bi-linear interpolation during training. 

Our work focuses on model architecture. We, therefore, majorly discuss UNET architecture and related architecture efforts such as UNET++. At same time, we believe that our proposed two-stream UNET networks can improve as well if they will be a part of nnUNET framework. Also, although we only discuss 2D segmentation in this paper, it is easy to extend our work to 3D cases.

% pre-trained models trained by nnUnet framework~\cite{isensee2021nnu} have improved the model performance. The nnUnet claims that basic Unet is hard to beat and improves training process in terms of data fingerprints. For the final results, nnUnet  focuses training data  

%% file: sections/twostream.tex
\section{GVF, two-Stream networks and image segmentation}
Two-stream networks demand for spatial and temporal components for learning spatial and temporal features. In this section, first of all, we introduce how two-stream networks are inspired by the two-stream hypothesis of human vision. Then we explain the relationship between GVF and pixel moving to the boundaries, and finally we discuss why two-steam networks can be used for are image segmentation. 
\begin{wrapfigure}[20]{r}{2.9in}
	\vspace{-10pt}
	\center
	\includegraphics[width=0.5\textwidth]{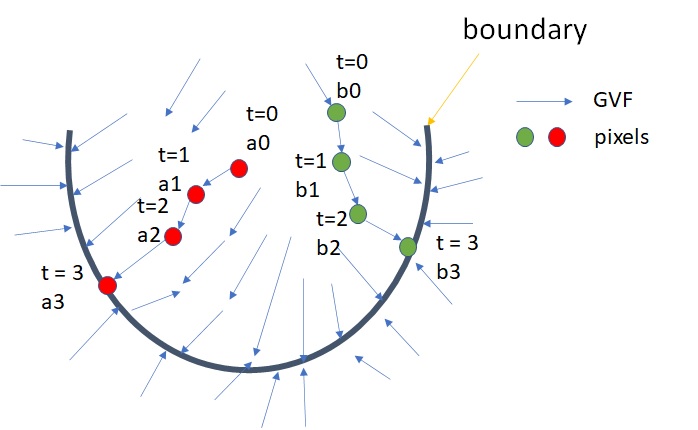}
	\caption{Visualization of relation between GVF and pixel motion analysis. The blue denotes GVF and red circle object is a pixel. Under the GVF, the object moves to edges step by step. For pixel a (red) will move from $a0\rightarrow  a1 \rightarrow a2\rightarrow a3$ at time $t0,t1,t2,t3$, and pixel b (green) will move from $b0\rightarrow  b1 \rightarrow b2\rightarrow b3$ at time $t0,t1,t2,t3$,. $a3$ and $b3$ are pixels of the boundary.}
	\vspace{-4pt}
	\label{fig:gvf}
\end{wrapfigure}
\label{sec_twostream}
\subsection{Two-stream hypothesis of human vision and action recognition}
In terms of two-streams hypothesis~\cite{goodale1992separate,ungerleider1994and}, the human visual cortex contains two streams~(pathways): a ventral stream and a dorsal stream. The ventral stream reasons about object identity~(object recognition), and the dorsal stream reasons about spatial relationships without regard for semantics~(motion analysis). Therefore, in action recognition~\cite{simonyan2014two,feichtenhofer2016convolutional} and other tasks in videos such as semantic robotic grasping~\cite{jang2017end}, one stream is trained to learn spatial features and the other stream is trained to learn temporal features.      
% pathways: the ventral stream for object recognition
% and the dorsal stream for motion recognition.

For two-stream networks for action recognition in videos, two inputs of the networks are naturally from RGB value of each frame and optical flow computed from adjacent flames~\cite{horn1981determining,sun2010secrets}. It is clear that the spatial steam acting as the ventral stream for learning spatial information from RGB data of each frames and temporal stream acting as dorsal stream for learning motion patterns in videos from optical flow.

For two-stream networks for image segmentation, the questions are \textit{\textbf{what can be fed into two-stream network for be trained and why two-stream networks for segmentation task?}}  The following two subsections will answer them. 
% of two-stream networks for image segmentation is natural. It is RGB value of the image as well. However, it is difficult for the field stream input to find which low-level features. In this section, first of all, we introduce gradient vector flow and explain why GVF field is well-suited for our goal. Then we detail our proposed two-stream networks. 

\subsection{GVF and two-stream networks}
\label{subsec_gvf}
\paragraph{Background of GVF.} GVF is the vector field that is produced by a process that smooths and diffuses an input vector field. It is usually used to create a vector field from images that points to object edges~\footnote{In image processing and computer vision, there are small differences among definitions of edge, contour, boundary. In this paper, we don't strictly distinguish among them. If don't mention, all of them mean the boundary of an object.} from a distance~\cite{xu1998snakes}. The GVF is given by $\mathbf{V}(x,y) = [u(x,y), v(x,y)]$ at position $(x,y)$ of an image. In this paper, we focus on CNNs and image segmentation, and more details of the GVF's mathematical theory and computational solutions can be seen in~\cite{xu1997gradient,xu1998snakes}.

\paragraph{Why GVF is a well-suitable for two-steam architecture?}

% In terms of two-streams hypothesis~\cite{goodale1992separate}, the human visual cortex contains two pathways: the ventral stream for object recognition and the dorsal stream for motion recognition. If a pixel is considered as an object, the GVF can seen as the force to push the object moving to edge. Like understanding object motion through optical flow, analyzing GVF field can understand pixels moving to edges.

According to newton's second law, motion patterns associate with forces that push the objects along with the direction of the force. We consider each pixel of an image is a component of an object. As shown in Figure.~\ref{fig:gvf}, pixels will move from the current position to the boundaries of the objects step by step under the GVF. The objective of image segmentation is to find the boundaries of the objects. As a result, GVF patterns represent how components move to the boundary of the object. Consequently, features from GVF infer to the boundaries in images. 

For example, as shown in Figure.~\ref{fig:gvf}, given two objects, A~(red) and B~(green), we can imagine that the ``A" object  moves from position a0, a1, a2, a3 and stop in a2 because a3 is located in boundary. The same as object ``B".

\subsection{Two-stream networks and image segmentation}
If we consider semantic segmentation as pixel moving to boundary task, the VS is trained to learn how pixels move to the object boundary, and the SS is train to learn to recognize objects. It is obvious that it is an exact process of the image semantic segmentation task. As a consequence, two-stream networks are suitable to image segmentation. 

There are two major reasons that two-stream networks are well-suited for medical image segmentation: (1) Each objects~(organs) in medical images have their-owned shape. and (2) The relationship among the location of the objects are fixed. There are some good patterns in GVF that offer visual features to be learned in VS for segmenting.   

%% file: sections/tech_approach.tex
\section{Methodology}
We now present our approach for semantic segmentation in medical images. Specifically, we first address the basic architecture of our spatial and vector field feature extractors. Then we detail our fusion method with the convolution. Finally, we discuss our idea of losses function in our training.   
\label{sec_techapproach}
\subsection{Spatial and vector field feature extractor}
\label{subsec_featureextractor}
UNET has shown that it is is one of the powerful CNNs for learning visual representation in medical tasks. In our two-stream networks, we use the same network architecture excluding the last output layer of basic UNET developed in~\cite{ronneberger2015u} as our spatial and vector field feature extractors, respectively. More details of UNET architecture can be seen in~\cite{ronneberger2015u}. The total of UNET has 23 convolutional layers including final convolutional layer with $1\times 1$ convolution. Both steams in our architecture  have 22 convolutional layers and the channels of outputs are 64. In the kernel size of both extractors are $3\times 3$, stride is 1.   
% The UNET consists of the repeated application of two 3x3 convolutions (unpadded convolutions), each followed by a rectified linear unit (ReLU) and a 2x2 max pooling operation with stride 2 for downsampling. At each downsampling step we double the number of featurechannels. Every step in the expansive path consists of an upsampling of thefeature map followed by a 2x2 convolution (“up-convolution”) that halves thenumber of feature channels, a concatenation with the correspondingly croppedfeature map from the contracting path, and two 3x3 convolutions, each fol-lowed by a ReLU. The cropping is necessary due to the loss of border pixelsin every convolution. At the final layer a 1x1 convolution is used to map each64-component feature vector to the desired number of classes. In total the net-work has 23 convolutional layers
\subsection{Two-stream Networks}
In this subsection, we extend the basic UNET to two-steam UNET networks by adding a similar architecture but with GVF as the input. Figure~\ref{fig:twostream} shows the main architecture of our proposed two-stream UNET networks.

There are four major techniques to fuse multiple feature maps~\cite{xing2019coupling}: sum fusion, max fusion, concatenation fusion, and convolution fusion. In our experiments, all of them can be used in our two-stream networks and improve the performances. In average, convolution fusion is better than others. Therefore, we use convolution fusion in this paper. 

Given two feature maps, spatial feature maps $\mathbf{X}^{s}$ and vector field features, $\mathbf{X}^{v}$, and they are concatenated firstly by:
\begin{equation}
\label{eq_cat}
y^{cat} = f^{cat}(\mathbf{X}^{s}, \mathbf{X}^{v})
\end{equation}
where $f^{cat}$ is a concatenation operator and stacks two features maps along the channel dimension.\\
Following it, a learnable of filter $\mathbf{f} \in \mathbb{R}^{M\times N}$ and biases $b \in \mathbb{R}^{N} $ is appended:
\begin{equation}
\label{eq_fusion}
y^{conv} = y^{cat} * \mathbf{f} + b
\end{equation}
where $*$ denotes convolution, $M$ denotes the sum number of channels of spatial and vector streams and $N$ denotes the number of output channels. Filter $\mathbf{f}$ provides a flexible approach to project the channel dimension from $M$ to $N$ with weighted combining of outputs of the two streams.  

\subsection{Complete loss functions}
%...

% This is my link:  \url{https://github.com/Christophe-Foyer/GVF-Python/}\\
In order to balance contributions of different losses during training, we train our networks with a weighed sum of dice and binary cross-entropy loss:
\begin{equation}
\label{eq_loss}
\mathbf{\mathit{L_{total}}} = \alpha \mathbf{\mathit{L_{dice}}} + (1-\alpha) \mathbf{\mathit{L_{bce}}}
\end{equation}
where $\alpha \in [0\:1]$, and was estimated by 

\begin{equation}
%\vspace{-1}
\label{eq_weight}
\alpha \propto \frac{\nabla \mathbf{\mathit{L_{bce}}}}{\nabla \mathbf{\mathit{L_{bce}}} + \bigtriangledown \mathbf{\mathit{L_{dice}}}}
\end{equation}
where $\nabla$ is a gradient operator. In our experiments, we calculate gradient of losses in two epochs. 

%% file: sections/experiment.tex
\section{Experiment}
\label{sec_expr}
\subsection{Dataset}
\label{subsec_dataset}
This section details the three widely-used datasets that we evaluate our approach in this paper and our experiment settings. 

% \begin{wrapfigure}[18]{r}{2.6in}
% \vspace{-25pt}
% \center
% \includegraphics[width=0.45\textwidth]{figs/gvf}
% \caption{Visualization of relation between GVF and pixel motion analysis. The blue denotes GVF and red circle object is a pxiel. Under the GVF, the object moves to edges step by step. }
% \vspace{-25pt}
%  \label{fig:gvf}
% \end{wrapfigure}

% \begin{wraptable}[5]{r}{0.65\linewidth}
% \begin{table}[!t]
%     \vspace{-0.3cm}
  
%     \caption{\label{table:nuclei_deepsupervision} 
%     Comparison experiments Nuclei.
%     }
%     \begin{tabular}{c|c}
%          &  \\
%          & 
%     \end{tabular}
%     % \vspace{-0.1cm}
%             \setlength{\tabcolsep}{0.6mm}{\begin{tabular}{ccccc}
%     \toprule
%     Method  &UNET   &UNET++   &UNET++ DS  &Ours \\
%     \midrule
%     Dice &83.9&84.2&84.3&84.6 $\pm$0.01 \\
 
%     %\midrule
    
%     \bottomrule
%     \vspace{-0.2cm}
%     \end{tabular}}
% \end{talbe}
% % \end{wraptable} 
\paragraph{Electron Microscopy~(EM)~\cite{ewald2012mammary}} The dataset consists of 30 images ($512\times512$), and was EM segmentation challenge as a part of IEEE International Symposium on Biomedical Imaging~(ISBI) 2012. The labeled images are split into training (24 images), validation (3 images), and test (3 images) datasets.
%The dataset is pro-vided by the EM segmentation challenge [30] as a part ofISBI 2012. The dataset consists of 30 images (512×512pixels) from serial section transmission electron microscopyof the Drosophila firt instar larva ventral nerve cord (VNC).Referring to the example in Fig.3, each image comes with acorresponding fully annotated ground truth segmentation mapfor cells (white) and membranes (black). The labeled images are split into training (24 images), validation (3 images), andtest (3 images) datasets. Both training and inference are donebased on 96×96 patches, which are chosen to overlap by halfof the patch size via sliding windows. Specifically, during theinference, we aggregate predictions across patches by votingin the overlapping areas. In our work, both training and inference are done based on ($96\times96$). Specifically, during the inference, unlike work in~\cite{ronneberger2015u,zhou2019unetplusplus}, we don't use aggregate predictions across patches by voting in the overlapping areas. The size of patches are $96\times96$ without overlapping. It is reasonable to use pro-process techniques to improve results in medical applications. However, the objective of this paper is to investigate on model architecture. With much pro-processing, 

\paragraph{Automated cardiac diagnosis challenge~(ACDC)~\cite{acdc}}The ACDC challenge collects exams from different patients acquired from MRI scanners. 
%Cine MR images were acquired in breath hold, and a series of short-axis slices cover the heart from the base to the apex of the left ventricle, with a slice thickness of 5 to 8 mm. The short-axis in-plane spatial resolution goes from 0.83 to 1.75 mm2/pixel.
Each patient scan is manually annotated with ground truth for left ventricle~(LV), right ventricle~(RV) and myocardium (MYO). These MRI images are randomly split to of 70 training cases (1930 axial slices), 10 cases for validation and 20 for testing.

\paragraph{Nuclei~\cite{nuclei}} The dataset is from the Data Science Bowl 2018 segmentation challenge and consists of 670 segmented nuclei images from different modalities.  Images are randomly split into a training set (50\%), a validation set (20\%), and a test set (30\%).
%We then use a sliding window mechanism to extract 96×96patches from the images, with 32-pixel stride for training and validating model, and with 1-pixel stride for testing.
% \subsection{Experiment settings}
% \paragraph{Models and Datasets}

% \paragraph{Code and Datasets}
% ISBI 2012 chllenges, results compared to Unet, Unet++, already in paper\\
% BraTs 2017 code from \url{https://github.com/cv-lee/BraTs}

% EM dataset dice 0.3 BCE 0.7

% code \url{https://github.com/dootmaan/mt-unet}

% code \url{https://github.com/bermanmaxim/LovaszSoftmax}

% \url{https://github.com/4uiiurz1/pytorch-nested-unet}

% \url{https://github.com/milesial/Pytorch-UNet}

% \url{https://github.com/Christophe-Foyer/GVF-Python}
% parameter is (0.2, 80)

% mu is GVF regularization coefficient and intel is the number 80 

\subsection{Experiment setting}

% Pytorch 1.8, Python3.7, V100 GPU 32Gmemeory 20.10  Intel(R) Xeon(R) Platinum 8260M CPU @ 2.40GHz 791217908 kB

We implemented the two-stream UNET networks in PyTorch~1.8~\cite{pytorch18}. 
Implementation of UNET backbone is from~\footnote{\label{unet}\url{https://github.com/milesial/Pytorch-UNet}} and implementation of GVF is from~\footnote{\url{https://github.com/Christophe-Foyer/GVF-Python}}. In computation of GVF, the parameter mu, which is the GVF regularization coefficient, was set as $0.2$, and the number of iteration of computation was set as $80$. All our experiments were conducted on a system with a Nvidia Tesla V100 cards with $32G$ memory. Our models were trained on one GPU card. The operation system is Ubuntu $20.10$. 
% Pascal graphics card.
    %   and ITER is the number of iterations that will be computed. 
%and used the C3D features and optical flow
% as inputs. As discussed before, we used mixed (high and
% low resolution) data in the training stage, but no high resolu-
% tion information is used in testing. We used the root-mean-
% square propagation (RMSprop) [23] update rule to learn
% the network parameters with fixed learning rate 10 −3 and
% weight decay 0.0005. The whole training process stopped
% at 50 epochs, with the batch size set to 256. All our exper-
% iments were conducted on a system with a Nvidia Titan X
% Pascal graphics card.

%% file: sections/evaluation.tex
\section{Evaluation and results}
In this section, we first discuss different evaluation methods. Then we compared our results to some of SOTA mehtods on three popular datasets, EM, Nuclei and ACDC datasets. The images of these dataset are acquired from MRI, CT, and rightfield VS fluorescence. Our results are competitive with the SOTA. 
\label{sec_eval}
\subsection{Evaluation}
Most of previous works in EM and Nuclei datasets~\cite{ronneberger2015u,zhou2019unetplusplus}, the predictions are aggregated across patches by voting in the overlapping areas.For instance, both SOTA results EM dataset reported from~\cite{ronneberger2015u,zhou2019unetplusplus}, the images are extracted to $96\times96$ patches with half of patch size overlaps. The final results are the aggregation with different patches with voting rule. 

It is entirely reasonable to use post-process techniques to improve results and evaluate them for medical applications because medical image segmentation aims at aiding clinical applications. However, a big disadvantage of this type of evaluation is that the improvements of the approaches cannot figure out which parts~(models and post-processing) attribute to and how much improvements respectively. In particular for model architecture work, the evaluation on these predictions is not effective because of the model performance is not clear.

Although some of results of previous works listed in this paper are still with post-processing such as EM dataset, our predictions are the models predictions without post-processing step. Even so, our proposed two-steam UNET networks have achieved SOTA in three datasets.   

In this paper, we measure our method with either intersection-over-union~(IoU), also known as the Jaccard index, or Dice score~(DSC). In order to compare the previous works, which type of metrics was reported depends on that of the previous work results. 

\subsection{Experiment results}
% Since the results of the models listed in different papers are not exactly same, in order to avoid confusion, the results that we list in the table is from the latest paper~\cite{wang2022mixed}   

\paragraph{Results on Electron Microscopy~(EM) dataset}For training phase, the images~($512\times 512$) was cut to $96 \times 96$ patches with half of the patches size overlaps via sliding windows. For our inference phase, the images~($512\times 512$) was cut to $96 \times 96$ patches without overlaps. During training, the data augmentation methods that we took were image rotation, vertical and horizontal clips. The optimizer was ADAM optimizer, the learning rating is $0.0003$, and batch size is 64. The total training epochs are 90 with early stopping. The losses function is weighted BCE and dice losses in Eq~\ref{eq_loss}. The parameter $\alpha$ was set as $0.3$. We trained our models from scratch. The training codebase that we revised for our two-steam networks training was from~\ref{unet} . 

\begin{table}[!t]
\caption{\label{table:EM} 
    Semantic segmentation results measured by IOU (mean$\pm$s.d.\%) on EM dataset for UNET, UNET++ and our approach. The results of UNET and UNET++ were in~\cite{zhou2019unetplusplus}. Both of them are with aggregation with voting on overlaps. Our results are directly predicted from two-stream models without post-processing. The best results are highlighted with bold. The same as others, for our approach, we have performed 20 independent trials. 
    }
\centering
    %   \begin{tabular}{c|c}
    %      &  \\
    %      & 
    % \end{tabular}
    % \vspace{-0.1cm}
            \setlength{\tabcolsep}{10pt}{\begin{tabular}{cccc}
    \toprule
      &UNET~\cite{ronneberger2015u}   &UNET++~\cite{zhou2019unetplusplus}   &Ours \\
    \midrule
    IOU &86.83$\pm$0.43&88.92$\pm$0.14& \textbf{89.12 $\pm$0.18} \\
    %\midrule
    \bottomrule
    \vspace{-0.2cm}
    \end{tabular}}
\end{table}

\begin{table}[!tb]
\caption{\label{table:adac} 
    Semantic segmentation results measured by IOU on ADAC dataset. The results of others were in~\cite{wang2022mixed}. The top two methods in the Table~\ref{table:adac} are UNET variant models, and others are ViT-based methods. The best results are highlighted with bold. 
    %\footnote{Since the results of the models listed in different papers are not exactly same, in order to avoid confusion, the results that we list in the table is from the latest paper~\cite{wang2022mixed}} 
    %\footnote[2]{New Radio Cellular Networks}
    }
\centering
    %   \begin{tabular}{c|c}
    %      &  \\
    %      & 
    % \end{tabular}
    % \vspace{-0.1cm}
    \setlength{\tabcolsep}{10pt}{\begin{tabular}{ccccc}
    \toprule
      Method &Average DSC(\%)  & RV  &MYO &LV \\
    \midrule
    R50 UNET~\cite{chen2021transunet} &87.60&84.62&84.52&93.68\\
    R50 AttnUNET~\cite{chen2021transunet} &86.90&83.27&84.33&93.53\\
 
    \midrule
    ViT-CUP~\cite{wang2022mixed}&83.41&80.93&78.12&91.17\\
    R50 ViT~\cite{chen2021transunet}&86.19&82.51&83.01&93.05\\
    TransUNET~\cite{chen2021transunet}&89.71&86.67&87.27&95.18\\
    Swin-UNET\cite{chen2021transunet}&88.07&85.77&84.42&94.03\\
    MT-UNET\cite{wang2022mixed}&90.43&86.64&\textbf{89.04}&\textbf{95.62}\\
    \midrule
    Ours &\textbf{90.90}&\textbf{88.62}&88.72&95.37\\
    \bottomrule
    \vspace{-0.2cm}
    \end{tabular}}
\end{table}

As shown in Table~\ref{table:EM}, compared to two benchmarks, UNET and UNET++ at the same settings, our approach has achieved the best performance. It shows that two-stream UNET networks are effective architecture to learn features of spatial and GVF domains.

\paragraph{Results on ACDC dataset.} For both training and inference phases, the resolution of the input images is~$224\times 224$. During training, the data augmentation methods that we took were image rotation, vertical and horizontal clips. The optimizer was ADAM optimizer, the learning rating is $0.0003$, and batch size is 12. The total training epochs are 90 with early stopping. The losses function is weighted CE and dice losses. The parameter $\alpha$ was set as $0.5$. We trained our models from scratch. The training codebase that we revised for our two-steam networks training was from~\footnote{\url{https://github.com/dootmaan/mt-unet}}. 

As shown in Table~\ref{table:adac}, our approach has achieved top 1 results in average results of three categories and RV task, and ranks second in MYO and LV segmentation. Furthermore, ours outperform marginally other CNNs methods at all in all segmentation tasks. Since the results of the same models on ADAC datasets reported in different papers are not exactly same, in order to avoid confusion, the results that we list in the Table~\ref{table:adac} is from the latest paper~\cite{wang2022mixed}.  

\paragraph{Results Nuclei dataset}For both training and inference phases, the resolution of the input images is $96\times 96$. During training, the data augmentation methods that we took were image rotation, vertical and horizontal clips. The optimizer was ADAM optimizer, the learning rating is $0.0003$, and batch size is 16. The total training epochs are 90 with early stopping. The loss was lov{\'a}sz-softmax loss~\cite{berman2018lovasz}. The reason that we used the lov{\'a}sz-softmax loss that beside is designed for image segmentation and a better in the task, we also evuluated our two-steam networks with different looses. The parameter $\alpha$ was set as $0.5$. We trained our models from scratch. The training codebase that we revised for our two-steam networks training was from~\footnote{{\label{nu}}\url{https://github.com/4uiiurz1/pytorch-nested-unet}}. 

As shown in Table~\ref{table:bts}, ours are better than the results of UNET, UNET++ and UNET++ with deep supervision training. It points out that our tow-stream UNET networks are robust, meaning they can be successfully trained in variety of situations, for example ours outperform UNET and UNET++ on different kinds of losses. 

\subsection{Limitations}
As our proposed two-stream UNET networks are designed for medical image datasets, in which the size of dataset is tiny. We will apply the models to more big datasets to compare his performance to  deeper models such as ViT-based models. In medical images, the number of categories are small and the objects~(organs) have a fixed shape. What is more, the organs have the fixed position, meaning the location relationship among objects in medical images are fixed as well. These lead to GVF of medical images have good patterns. On the other hand, these properties of natural images are weaker. As a result, it will undermine the results of two-stream network for image segmentation in the wild.

\begin{table}[!t]
\caption{\label{table:bts} 
    emantic segmentation results measured by IOU (mean$\pm$s.d.\%) on Nuclei dataset for UNET, UNET++, UNET++DS  and our approach. NET++DS denotes UNET++ with deep supervision during traning The results of others were in~\ref{nu}, last access on May 1, 2022. The best results are highlighted with bold. For our approach, we have performed 10 independent trials.
    }
\centering
    %   \begin{tabular}{c|c}
    %      &  \\
    %      & 
    % \end{tabular}
    % \vspace{-0.1cm}
            \setlength{\tabcolsep}{10pt}{\begin{tabular}{ccccc}
    \toprule
      &UNET   &UNET++   &UNET++ DS  &Ours \\
    \midrule
    IOU(\%) &83.9&84.2&84.3&\textbf{84.6 $\pm$0.01} \\
    \bottomrule
    \vspace{-0.2cm}
    \end{tabular}}
\end{table}

%is with siding window and voting in the overlapping areas, meaning  oth training and inference are donebased on 96×96 patches, which are chosen to overlap by halfof the patch size via sliding windows. Specifically, during theinference, we aggregate predictions across patches by votingin the overlapping areas. In our work, both training and inference are done based on ($96\times96$). Specifically, during the inference, unlike work in~\cite{ronneberger2015u,zhou2019unetplusplus}, we don't use aggregate predictions across patches by voting in the overlapping areas. The size of patches are $96\times96$ without overlapping. It is reasonable to use pro-process techniques to improve results in medical applications. However, the objective of this paper is to investigate on model architecture. With much pro-processing, 

    %\footnote[2]{New Radio Cellular Networks

%% file: sections/conclusion.tex
\section{Conclusion and future work}
\label{sec_conclusion}
In this work we propose a novel two-stream UNET networks, which incorporates separate spatial and vector streams, for semantic segmentation in medical images.The backbone of both streams is UNET architecture. In perspective of motion, GVF indicates that the pixels of an images moves to the boundary of an object that the pixels belongs to. In term of motion analysis, the models can find the boundaries of the objects. As a consequence, two-stream networks are successfully trained on RGB input and GVF input that is calculated from RGB data. Extensive experiments validate the effectiveness of our proposed approach on semantic segmentation in medical images.
%Currently it appears that training a VS on GVF is significantly better than training on raw stacked frames\cite{ronneberger2015u}   

Aside from SOTA results on three challenging datasets, our approach also retains an attractive open scheme. There are two directions for future work are as followings:

\begin{enumerate}
   \item Combining with other existing modules such as attention module
   , which have shown that models with attention have improved model training in medical image segmentation\cite{chen2016attention,vaswani2017attention}. We believe that attention module will offer an effective way to improve two-stream UNET networks for medical image segmentation.   
    \item The other way is to combine with some data pre-cessing for improving training data quality, such as nnUNET. nnUNet has exhibited that the models benefited from better training data. Given CNNs are the data-driven method, high quality data will definitely improve the performance of CNNs models including our approach. 
    
\end{enumerate}